\documentclass{article}
\usepackage{spconf,amsmath,graphicx,hyperref,amsmath,amssymb}

\title{MMLatch: Bottom-up Top-down Fusion for Multimodal Sentiment Analysis}
\name{Georgios Paraskevopoulos$^{\dagger,\ast}$ \qquad Efthymios Georgiou$^{\dagger,\ast}$ \qquad Alexandros Potamianos$^{\dagger,\ddagger}$}

\address{
$^\dagger$ National Technical University of Athens, Athens, Greece\\
$\ast$ Institute for Language and Speech Processing, Athena Research Center, Athens, Greece \\
$^\ddagger$  Behavioral Signal Technologies, Los Angeles, CA, USA}

\begin{document}
%
\maketitle
\begin{abstract}
Current deep learning approaches for multimodal fusion rely on bottom-up fusion of high and mid-level latent modality representations (late/mid fusion) or low level sensory inputs (early fusion). 
Models of human perception highlight the importance of top-down fusion, where high-level representations affect the way sensory inputs are perceived, i.e. cognition affects perception. These top-down interactions are not captured in current deep learning models.
In this work we propose a neural architecture that captures top-down cross-modal interactions, using a feedback mechanism in the forward pass during network training.
The proposed mechanism extracts high-level representations for each modality and uses these representations to mask the sensory inputs, allowing the model to perform top-down feature masking.
We apply the proposed model for multimodal sentiment recognition on CMU-MOSEI.
Our method shows consistent improvements over the well established MulT and over our strong late fusion baseline, achieving state-of-the-art results.
\end{abstract}

\section{Introduction}
\label{sec:intro}

Multimodal processing aims to model interactions between inputs that come from different sources in real world tasks.
Multimodality  
can open ways to develop novel applications (e.g. Image Captioning, Visual Question Answering 
\cite{antol2015vqa, you2016image} etc.) 
or boost performance in traditionally unimodal applications (e.g. Machine Translation \cite{caglayan-etal-2019-probing}, Speech Recognition \cite{paraskevopoulos-etal-2020-multimodal, srinivasan-etal-2020-multimodal} etc.).
Moreover, modern advances in neuroscience and psychology hint that multi-sensory inputs are crucial for cognitive functions \cite{KLEMEN2012111}, even since infancy \cite{neil2006development}.
Thus, modeling and understanding multimodal interactions can open avenues to develop smarter agents, inspired by the human brain.

Feedback loops have been shown to exist in the human brain, e.g. in the case of vocal production \cite{houde2015cortical} or visual-motor coordination \cite{SHAFER2019112214}.
Human perception has been traditionally modelled as a linear (bottom-up) process (e.g. reflected light is captured by the eye, processed in the prefrontal visual cortex, then the posterior visual cortex etc.).
Recent studies have highlighted that this model may be too simplistic and that high level cognition may affect low-level visual \cite{bar2013top,TEUFEL201717} or audio \cite{sohoglu2012predictive} perception. 
For example, studies state that perception may be affected by an individual's long-term memory \cite{lupyan2017objective}, emotions \cite{balcetis2010wishful} and physical state \cite{proffitt1995perceiving}.
Researchers have also tried to identify brain circuits that allow for this interplay \cite{MANITA20151304}.
While scientists still debate on this subject \cite{firestone2014top}, such works offer strong motivation to explore if artificial neural networks can benefit from multimodal top-down modeling.


Early works on multimodal machine learning use binary decision trees \cite{lee2011emotion} and ensembles of Support Vector Machines \cite{rozgic2012ensemble}. Modeling contextual information is addressed in \cite{metallinou2012context, wollmer2013lstm, shenoy-sardana-2020-multilogue} using Recurrent Neural Networks (RNNs), while Poria et al. \cite{poria-cmkl-16} use Convolutional Neural Networks (CNNs). For a detailed review we refer to Baltruvsaitis et al. \cite{baltruvsaitis2018multimodal}. Later works use Kronecker product between late representations \cite{zadeh2017tensor, liu2018efficient}, while others investigate architectures with neural memory-like modules \cite{Zadeh_Liang_Poria_Vij_Cambria_Morency_2018, bagher-zadeh-etal-2018-multimodal}. Hierarchical attention mechanisms \cite{gu2018multimodal}, as well as hierarchical fusion \cite{georgiou2019deep} have been also proposed.
Pham et al. \cite{pham2019found} learn cyclic cross-modal mappings, Sun et al. \cite{sun2019multi} propose Deep Canonical Correlation Analysis (DCCA) for jointly learning representations. Multitask learning has been also investigated \cite{Khare_2020} in the multimodal context. Transformers \cite{vaswani2017attention} have been applied to and extended for multimodal tasks 
\cite{tsai-etal-2019-multimodal, vilbert2019, delbrouck-etal-2020-transformer, rahman-etal-2020-integrating}.
Wang et al. \cite{wang2019words} shift word representations based on non-verbal imformation. 
\cite{kumar2020gated} propose a fusion gating mechanism. \cite{tsai-etal-2020-multimodal} use capsule networks \cite{capsules2017} to weight input modalities and create distinct representations for input samples.

\begin{figure*}[h]
\centering
\includegraphics[width=.7\textwidth]{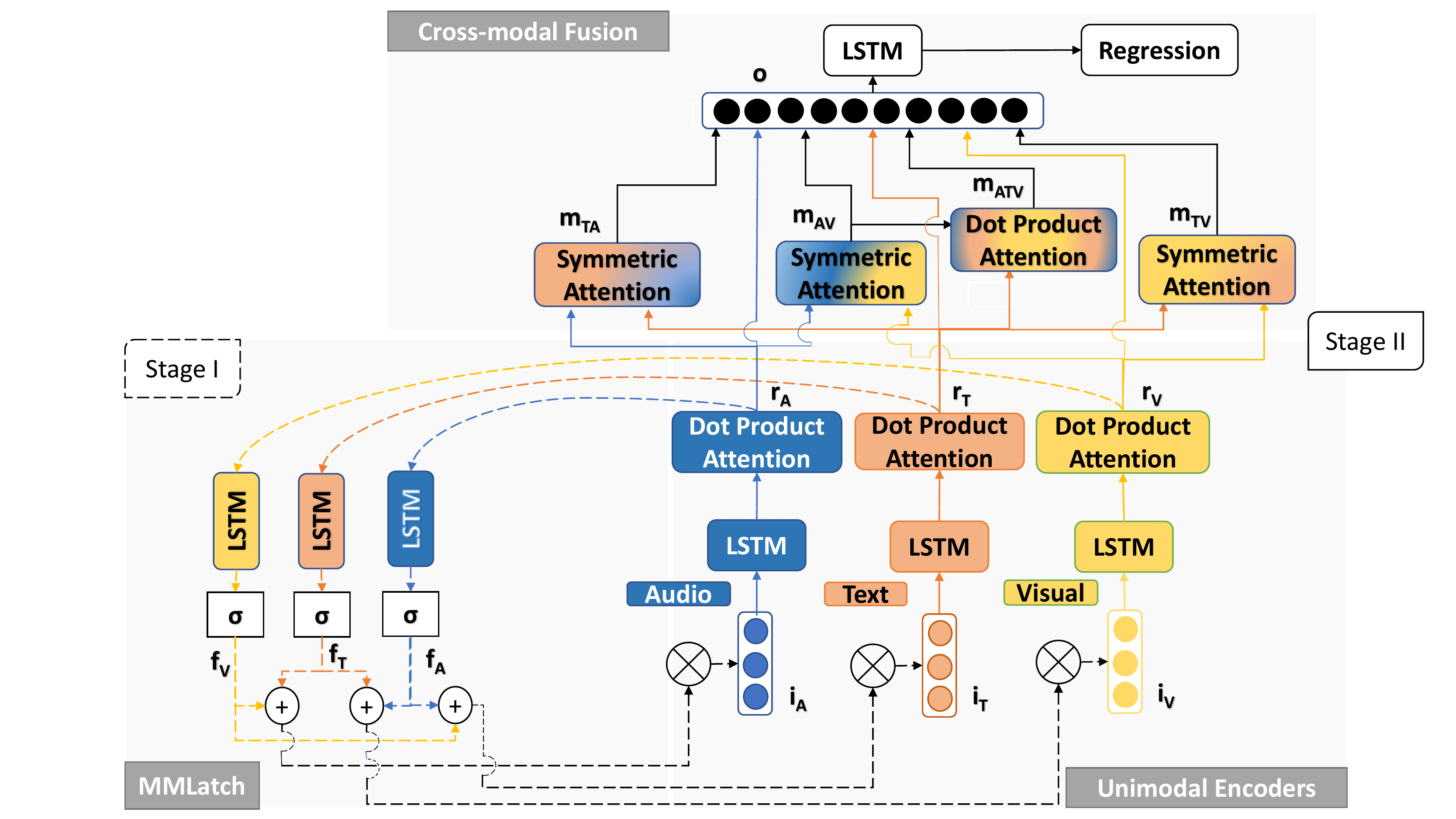}
\caption{Architecture overview of three high-level modules, composing the overall system: Unimodal encoders, Cross-modal fusion and MMLatch. Solid lines indicate the feedforward connections (bottom-up processing), while dashed lines indicate feedback connections (top-down processing). Colors indicate different modalities (Blue: Audio, Orange: Text, Yellow: Visual)}
\label{fig:arch}
\end{figure*}

In this work we propose MMLatch, a neural network module that uses representations from higher levels of the architecture to create top-down masks for the low level input features. The masks are created by a set of feedback connections. The module is integrated in a strong late fusion baseline based on LSTM \cite{hochreiter1997long} encoders and cross-modal attention.
Our key contribution is the modeling of interactions between high-level representations extracted by the network and low-level input features, using an end to end framework.
We integrate MMLatch with RNNs, but it can be adapted for other architectures (e.g. Transformers).
Incorporating top-down modeling shows consistent improvements over our strong baseline, yielding state-of-the-art results for sentiment analysis on CMU-MOSEI.
Qualitative analysis of learned top-down masks can add interpretability in multimodal architectures.
Our code will be made available as open source.

\section{Proposed Method}
\label{sec:proposed}

Fig.~\ref{fig:arch} illustrates an overview of the system architecture.
The baseline system consists of a set of unimodal encoders and a cross-modal attention fusion network, that extracts fused feature vectors for regression on the sentiment values. 
We integrate top-down information by augmenting the baseline system with a set of feedback connections that create cross-modal, top-down feature masks. 

\noindent\textbf{Unimodal Encoders:} 
Input features $i_A, i_T, i_V$ for each modality are encoded using three LSTMs. 
The hidden states of each LSTM are then passed through a Dot Product self-attention mechanism 
to produce the unimodal representations $r_A, r_T, r_V$,  where $A,T,V$ are the audio, text and visual modalities respectively.

\noindent\textbf{Cross-modal Fusion:}
The encoded unimodal representations are fed into a cross-modal fusion network, that uses a set of attention mechanisms to capture cross-modal interactions.
The core component of this subsystem is the symmetric attention mechanism, inspired by Lu et al. \cite{vilbert2019}.
If we consider modality indicators $k,l \in \{A, V, T\}, \, k \neq l$, $r_k, r_l \in \mathbb{R}^{B \times N \times  d}$ the input modality representations, we can construct keys $K_l = W^K_l r_l$, queries $Q_k = W^Q_k r_k$ and values $V_l = W^V_l r_l$ using  learnable projection matrices $W_{\{k,l\}}^{\{K,Q,V\}}$, and we can define a cross-modal attention layer as:


\begin{equation}
    a_{kl} =  s \left(  \frac{K_l^T Q_k}{\sqrt{d}} \right) V_l + r_k,
\label{eq:crossatt}
\end{equation}

\noindent where $s$ is the softmax operation and $B, \, N, \, d$ are the batch size, sequence length and hidden size respectively.
For the symmetric attention we sum the two cross-modal attentions:

\begin{equation}
m_{kl} = a_{kl} + a_{lk},
\label{eq:symatt}
\end{equation}


In the fusion subsystem we use three symmetric attention mechanisms to produce $m_{TA}$, $m_{TV}$ and $m_{AV}$. 
Additionally we create $a_{AVT}$ using a cross-modal attention mechanism (Eq.~\eqref{eq:crossatt}) with inputs $m_{AV}$ and $r_T$. 
These crossmodal representations are concatenated ($\mathbin\Vert$), along with the unimodal representations $m_A, m_T, m_V$ to produce the fused feature vector $o \in \mathbb{R}^{B \times N \times  7d}$ in Eq.~\eqref{eq:concat}.

\begin{equation}
    o = r_A \mathbin\Vert r_T \mathbin\Vert r_V \mathbin\Vert a_{AVT} \mathbin\Vert m_{AV} \mathbin\Vert m_{TV} \mathbin\Vert m_{TA}
\label{eq:concat}
\end{equation}

We then feed $o$ into a LSTM and the last hidden state is used for regression.
The baseline system consists of the unimodal encoders followed by the cross-modal fusion network.

\begin{table*}[h]
    \centering
    \small
    \begin{tabular}{|c|c|c|c|c|c|c|}
      \hline
       Model / Metric & Acc@7  & Acc@2 & F1@2  & MAE & Corr \\\hline\hline
       RAVEN \cite{wang2019words} $^*$ & $50.0$ &  $79.1$ & $79.5$ & $0.614$ & $0.662$ \\
MCTN \cite{pham2019found} $^*$ & $49.6$ &  $79.8$ & $80.6$ & $0.609$ & $0.670$ \\
Multimodal Routing \cite{tsai-etal-2020-multimodal} & $51.6$ &  $81.7$ & $81.8$ & - & - \\
MulT \cite{tsai-etal-2019-multimodal} & $51.8$ &  $82.5$ & $82.3$ & $\mathbf{0.580}$ & $\mathbf{0.703}$ \\ \hline
Baseline (ours) & $51.3 \pm 0.7$ & $81.9 \pm 0.7$ & $82.2 \pm 0.6$ & $0.593 \pm 0.005$ & $0.695 \pm 0.004$ \\
Baseline + MMLatch average (ours) & $\mathbf{52.0} \pm 0.2$ & $82.4 \pm 0.3$ & $\mathbf{82.5} \pm 0.3$ & $0.585 \pm 0.002$ & $0.700 \pm 0.004$\\
Baseline + MMLatch best (ours) & $\mathbf{52.1}$ & $\mathbf{82.8}$ &  $\mathbf{82.9}$ & $0.582$ & $\mathbf{0.704}$ \\  \hline
    \end{tabular}
    \caption{Results on CMU-MOSEI for MMLatch. Models indicated with $^*$ are reproduced for CMU-MOSEI by Tsai et al. \cite{tsai-etal-2019-multimodal}. In row ``MMLatch average'' we include  results averaged over five runs. Since other works do not report standard deviation, we also include row ``MMLatch best'', where we report the best of the five runs (lowest error).}
    \label{tab:my_label}
\end{table*}

\noindent\textbf{Top-down fusion:} 
We integrate top-down information by augmenting the baseline system with MMLatch, i.e. a set of feedback connections composing of three LSTMs followed by sigmoid activations $\sigma$.
The inputs to these LSTMs are $r_A, r_T, r_V$ as they come out of the unimodal encoders. Feedback LSTMs produce hidden states $h_A, h_T, h_V$.
The feedback masks $f_A, f_T, f_V$ are produced by applying a sigmoid activation on the hidden states $f_k = \sigma(h_k), \, k \in \{A, T, V\}$ and then applied to the input features $i_A, i_T, i_V$ using element-wise multiplication $\odot$, as:


\begin{equation}
    \widetilde{i}_k = \frac{1}{2}(f_j + f_l) \odot i_k 
    \label{eq:featselect}
\end{equation}
where $j, k,l \in \{A,V,T\}, \, k \neq l \neq m$.

 Eq.~\eqref{eq:featselect} describes how the feedback masks for two modalities are applied to the input features of the third.
For example, consider the case where we mask visual input features using the (halved) sum of text and audio feedback masks.
If a visual feature is important for both audio and text the value of the resulting mask will be close to $1$. If it is important for only one other modality the value will be close to $0.5$, while if it is irrelevant for text and audio the value will be close to $0$.
Thus, a feature is enhanced or attenuated based on it's overall importance for cross-modal representations.

This pipeline is implemented as a two-stage  computation.
During the first stage we use the unimodal encoders and MMLatch to produce the feedback masks $f_A, f_T, f_V$ and apply them to the input features using Eq.~\eqref{eq:featselect}.
During the second stage we pass the masked features $\widetilde{i}_A, \widetilde{i}_T, \widetilde{i}_V$ through the unimodal encoders and the cross-modal fusion module and use the fused representations for regression.


 



\section{Experimental setup}
\label{sec:setup}
We use CMU-MOSEI sentiment analysis dataset \cite{bagher-zadeh-etal-2018-multimodal} for our experiments.
The dataset contains $23,454$ YouTube video clips of movie reviews accompanied by human annotations for sentiment scores from -3 (strongly negative) to 3 (strongly positive) and emotion annotations.
Audio sequences are sampled at $20Hz$ and then $74$ COVAREP features 
are extracted.
Visual sequences are sampled at $15Hz$ and represented using Facet 
features.
Video transcriptions are segmented in words and  represented using GloVe. 
All sequences are word-aligned using P2FA. 
Standard train, validation and test splits are provided.

For all our experiments we use bidirectional LSTMs with hidden size $100$.
LSTMs are bidirectional and forward and backward passes are summed.
All projection sizes for the attention modules are set to $100$. 
We use dropout $0.2$.
We use Adam \cite{adam14} with learning rate $0.0005$ and halve the learning rate if the validation loss does not decrease for $2$ epochs.
We use early stopping on the validation loss (patience $10$ epochs).
During Stage I of each training step we disable gradients for the unimodal encoders.
Models are trained for regression on sentiment values using Mean Absolute Error (MAE) loss.
We use standard evaluation metrics: $7$-class, $5$-class accuracy (i.e. classification in $\mathbb{Z}\cap[-3, 3]$, $\mathbb{Z}\cap[-2, 2]$), binary accuracy and F1-score (negative in $[-3,0)$, positive in $(0, 3]$), MAE and correlation between model and human predictions.
For fair comparison we compare with methods in the literature that use Glove text features, COVAREP audio features and FACET visual features.




\begin{table*}[h]
    \centering
    \small
    \begin{tabular}{|c|c|c|c|c|c|c|c|}
      \hline
     Multimodal Encoder & Feedback Type & Acc@7  & Acc@2 & F1@2  & MAE & Corr \\\hline\hline
Baseline & - &  $51.3 \pm 0.7$ & $81.9 \pm 0.7$ & $82.2 \pm 0.6$ & $0.593 \pm 0.005$ & $0.695 \pm 0.004$ \\
Baseline & MMlatch (no LSTM) & $51.48 \pm 0.41$ & $82.07 \pm 0.47$ & $82.29 \pm 0.39$ & $0.592 \pm 0.002$ & $0.692 \pm 0.003$ \\ 
Baseline & MMLatch & $\mathbf{52.0} \pm 0.2$ & $\mathbf{82.4} \pm 0.3$ & $\mathbf{82.5} \pm 0.3$ & $\mathbf{0.585} \pm 0.002$ & $\mathbf{0.700} \pm 0.004$\\ \hline
MulT$^\dagger$ & - & $47.91 \pm 1.13$ &  $80.35 \pm 0.36$ & $80.54 \pm 0.52$ & $0.643 \pm 0.01$ & $0.648 \pm 0.02$ \\ 
MulT$^\dagger$ & MMLatch & $\mathbf{49.04} \pm 0.45$ & $\mathbf{80.65} \pm 0.43$ &  $\mathbf{81.07} \pm 0.38$ & $\mathbf{0.627} \pm 0.004$ & $\mathbf{0.665} \pm 0.003$ \\  \hline
    \end{tabular}
    \caption{Results on CMU-MOSEI when combining top-down feedback with different multimodal encoder networks.  MulT with $^\dagger$ is reproduced by us. We report results, averaged over five runs, along with standard deviations.}
    \label{tab:mult}
\end{table*}

\begin{figure*}[h]
\centering
\includegraphics[width=.7\textwidth]{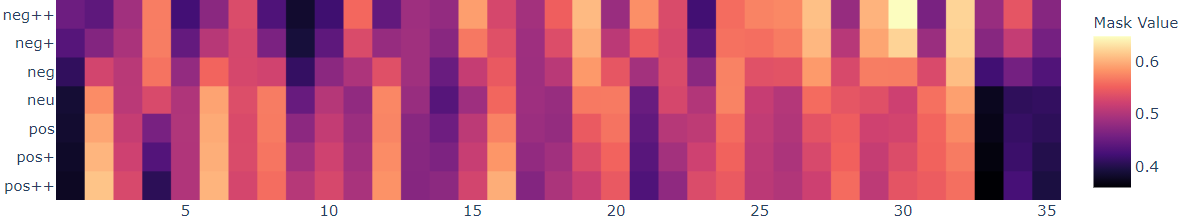}

\caption{Averaged top-down mask values for Facet features over all test samples across seven sentiment classes. neg++ indicates a sentiment score $\approx -3$, neg+ $\approx -2$, neg $\approx -1$, neu $\approx 0$, pos $\approx 1$, pos+ $\approx 2$ and pos++ $\approx 3$. }
\label{fig:heatmap}
\end{figure*}
\section{Experiments}
\label{sec:exp}

Table ~\ref{tab:my_label} shows the results for sentiment analysis on CMU-MOSEI.
The Baseline row refers to our late-fusion baseline described in Section~\ref{sec:proposed}, which achieves competitive to the state-of-the-art performance. 
Incorporating MMLatch into the baseline constistently improves performance and specifically, almost $1.0\%$ over the binary accuracy and $0.8\%$ over the seven class accuracy. 
Moreover, we observe lower deviation, w.r.t. the baseline, across experiments, indicating that top-down feedback can stabilize training.
Compared to state-of-the-art we achieve better performance for 7-class accuracy and binary F1 metrics in our five run experiments.
Since, prior works do not report average results over multiple runs so we also report results for the best (mean absolute error) out of five runs in the last row of Table~\ref{tab:my_label}, showing improvements across metrics over the best runs of the other methods.

In Table~\ref{tab:mult} we evaluate MMLatch with different multimodal encoders and different feedback types. The first three rows show the effect of using different feedback types. Specifically, first row shows our baseline performance (no feedback). For the second row we add feedback connections, but instead of using LSTMs in the feedback loop (Stage I in Fig.~\ref{fig:arch}), we use a simple feed-forward layer. The last row shows performance when we include LSTMs in the feedback loop. We observe that, while the inclusion of top-down feedback, using a simple projection layer results to a small performance boost, when we include an LSTM in the feedback loop we get significant improvements. This shows that choosing an appropriate mapping from high-level representations to low-level features in the feedback loop is important.

For the last two rows of Table~\ref{tab:mult} we integrate MMLatch with MulT architecture\footnote{We use the original code in this \href{https://github.com/yaohungt/Multimodal-Transformer}{GitHub Link}} \cite{tsai-etal-2019-multimodal}. Specifically, we use MMLatch, as shown in Fig.~\ref{fig:arch} and swap the baseline architecture (unimodal encoders and cross-modal fusion) with MulT. We use a $4$-layer Transformer model with the same hyperparameter set and feature set described in the original paper \cite{tsai-etal-2019-multimodal}.
The output of the fourth (final) layer is used by MMLatch to mask the input features.
First, we notice a performance gap between our reproduced results and the ones reported in the original paper (fourth row of Table~\ref{tab:mult}). Other works \cite{wen2021cross,sourav-ouyang-2021-lightweight} have reported similar observations.
We observe that the integration of MMLatch with MulT yields significant performance improvements across metrics.
Furthermore, similarly to Table~\ref{tab:my_label}, we observe that the inclusion of MMLatch reduces standard deviation across metrics. 
Overall, we observe that the inclusion of MMLatch results to performance improvements for both our baseline model and MulT with no additional tuning, indicating that top-down feedback can provide stronger multimodal representations.


Fig.~\ref{fig:heatmap} shows a heatmap of the average mask values $\frac{1}{2}(f_T + f_A)$.
This mask is applied to the input visual features $i_V$, i.e. $35$ Facet features. The average mask values range from $0.36$ to $0.65$ and depicted across $7$ sentiment classes.
Some features are attenuated or enhanced across all classes (e.g. features $1$ or $32$). 
Interestingly, some features are attenuated for some classes and enhanced for others (e.g. feature $2$).
More importantly this transition is smooth, i.e. mask values change almost monotonically as the sentiment value increases from $-3$ to $+3$, indicating welll-behaved training of MMlatch. We observe the same for Covarep masks.

\section{Conclusions}
\label{sec:conclusions}

We introduce MMLatch, a feedback module that allows modeling top-down cross-modal interactions between higher and lower levels of the architecture. 
MMLatch is motivated by recent advances in cognitive science, analyzing how cognition affects perception and is implemented as a plug and play framework that can be adapted for modern neural architectures.
MMLatch improves model performance over our proposed baseline and over MulT. The combination of MMLatch with our baseline achieves state-of-the-art results.
We believe top-down cross-modal modeling can augment traditional bottom-up pipelines, improve performance in multimodal tasks and inspire novel multimodal architectures.

In this work, we implement top-down cross-modal modeling as an adaptive feature masking mechanism. 
In the future, we plan to explore more elaborate implementations that directly affect the state of the network modules from different levels in the network.
Furthermore, we aim to extend MMLatch to more tasks, diverse architectures (e.g. Transformers) and for unimodal architectures.
Finally, we want to explore the applications top-down masks for model interpretability.



\section{References}
\small
\bibliographystyle{IEEEbib}
\bibliography{refs}

\end{document}